\documentclass[sigconf]{acmart}

\acmConference[Conference acronym 'XX]{Make sure to enter the correct conference title from your rights confirmation email}{Conference dates}{Woodstock, NY}
\acmDOI{XXXXXXX.XXXXXXX}

\AtBeginDocument{%
  }

\usepackage{amsmath}
\usepackage{graphicx}
\usepackage{array}
\usepackage{tabularx}
\usepackage{multirow}
\usepackage{placeins}
\usepackage{float}
\usepackage{tikz}
\usetikzlibrary{backgrounds,calc}
\newfloat{template}{p}{lop}
\floatname{template}{Template}
\graphicspath{{figures/}}

\title[Learning Evidence-Grounded Reasoning for Context-Aware Time Series Forecasting]{Learning Evidence-Grounded Reasoning for Context-Aware \\Time Series Forecasting}
\author{Xingpeng Gao}
\email{gaoxingpeng@mail.ustc.edu.cn}
\affiliation{%
  \institution{State Key Laboratory of Cognitive Intelligence, University of Science and Technology of China}
  \city{Hefei}
  \country{China}}

\author{Mingyue Cheng}
\authornote{Corresponding author.}
\email{mycheng@ustc.edu.cn}
\affiliation{%
  \institution{State Key Laboratory of Cognitive Intelligence, University of Science and Technology of China}
  \city{Hefei}
  \country{China}}

\author{Daoyu Wang}
\email{wdy030428@mail.ustc.edu.cn}
\affiliation{%
  \institution{State Key Laboratory of Cognitive Intelligence, University of Science and Technology of China}
  \city{Hefei}
  \country{China}}

\author{Huajian Zhang}
\email{zhjustc@mail.ustc.edu.cn}
\affiliation{%
  \institution{State Key Laboratory of Cognitive Intelligence, University of Science and Technology of China}
  \city{Hefei}
  \country{China}}

\author{Xiaoyu Tao}
\email{txytiny@mail.ustc.edu.cn}
\affiliation{%
  \institution{State Key Laboratory of Cognitive Intelligence, University of Science and Technology of China}
  \city{Hefei}
  \country{China}}

\makeatletter
\renewcommand{\@mkauthors@iii}{%
  \gdef\and{}%
  \gdef\@currentauthors{}%
  \gdef\@currentaffiliation{}%
  \global\setbox\mktitle@bx=\vbox{%
    \noindent\unvbox\mktitle@bx\par\medskip
    \centering
    {\@authorfont
      Xingpeng Gao, Mingyue Cheng\footnotemark[1], Daoyu Wang,
      Huajian Zhang, Xiaoyu Tao\par}
    {\@affiliationfont
      State Key Laboratory of Cognitive Intelligence, University of Science and Technology of China\par
      Hefei, Anhui, China\par
      \{gaoxingpeng,wdy030428,zhjustc,txytiny\}@mail.ustc.edu.cn,
      \{mycheng\}@ustc.edu.cn\par}
    \bigskip}}
\makeatother

\begin{document}

\begin{abstract}
LLM-driven time series forecasting has become increasingly promising by leveraging contextual features to generate future values directly or refine forecast priors. However, incorporating contextual features alone does not
ensure that models can identify which features matter for forecasting and understand what they suggest about future dynamics.
We argue that the fundamental challenge is how to extract predictive evidence from contextual features and when and how to leverage it to guide forecasting decisions.
In this paper, we propose CastReasoner, an evidence-grounded reasoning model for context-aware time series forecasting.
Specifically, CastReasoner uses a structured chain of thought (CoT) to extract predictive evidence from contextual features, infer the future dynamics suggested by this evidence, and assess these dynamics against the forecast prior to guide adaptive forecasting decisions.
To enable this capability, we design a reasoning-centric two-stage post-training framework including supervised fine-tuning (SFT) and reinforcement learning (RL).
First, we synthesize reasoning trajectories through controllable multi-teacher guidance and select high-quality ones using a novel student-aware difficulty metric.
Second, we design fine-grained multi-objective rewards to further improve reasoning and forecasting.
To address heterogeneous learning needs across domains in both stages, we introduce a convergence-guided data balancing strategy. Through this two-stage training, CastReasoner improves both reasoning quality and predictive accuracy.
Extensive experiments across multi-domain forecasting benchmarks demonstrate that CastReasoner achieves superior forecasting performance.\footnote{Our code is available at \url{https://github.com/gaoxp112/CastMind-Public}.}
\end{abstract}

\maketitle

\begin{figure}[!t]
  \centering
  \includegraphics[width=\columnwidth]{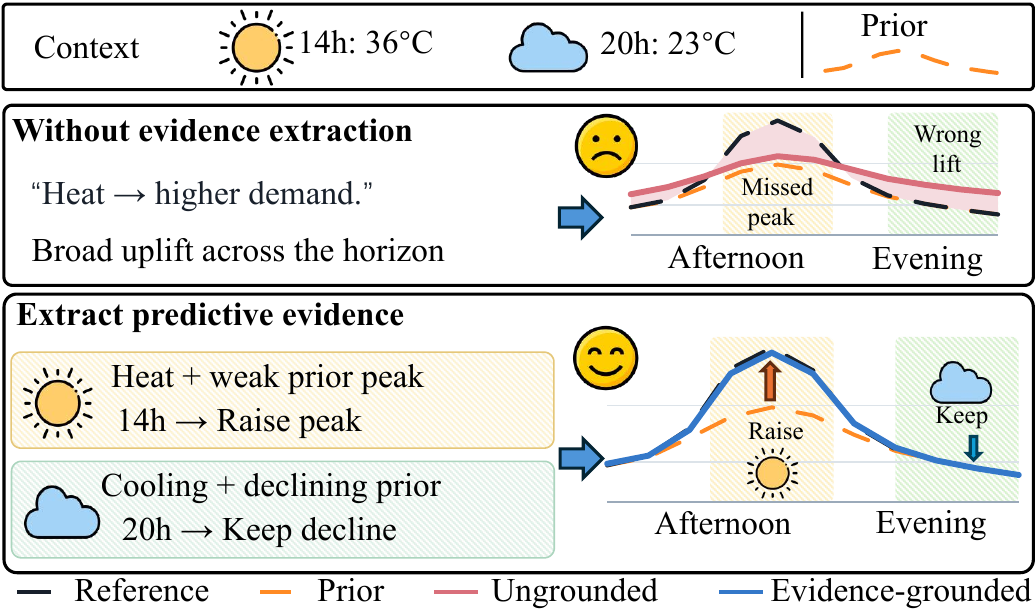}
  \caption{Illustrative demand forecasting with and without evidence extraction. Context and a forecast prior support raising the afternoon peak and retaining the evening decline.}
  \Description{The same context and forecast prior support two illustrative forecasting paths. Without extracting predictive evidence, a broad uplift misses the afternoon peak and overestimates evening demand. Evidence from afternoon heat and a weak prior peak supports raising the peak, while evening cooling and a declining prior support retaining the decline.}
  \label{fig:castreasoner-motivation}
\end{figure}

\section{Introduction}
Context-aware time series forecasting supports decisions in Web-connected systems, including Internet of Things (IoT) sensing~\cite{cai2025mate}, network traffic monitoring~\cite{lian2025undiff}, and urban mobility~\cite{hong2025stkopt}. By incorporating exogenous variables, calendar information, and domain knowledge, it accounts for external influences that historical observations do not fully reveal~\cite{huang2025exogenous,jia2024gpt4mts}. Recent LLM reasoning approaches increasingly incorporate contextual features into time series forecasting, attracting growing research interest~\cite{cheng2026timereasoner,zhou2026timer1}.

Existing LLM reasoning approaches for context-aware time series forecasting follow two routes. The first directly generates future values from historical patterns and contextual influences~\cite{cheng2026timereasoner,zhou2026timer1}. The second refines a forecast prior from a separate forecaster~\cite{liu2026posttime}, such as a time series foundation model pretrained on temporal observations to capture transferable patterns across series~\cite{ansari2024chronos,woo2024moirai}. An LLM assesses the prior against the context to determine appropriate adjustments. However, incorporating contextual features alone does not ensure that existing approaches identify predictive evidence or understand its implications for future dynamics. Predictive evidence refers to cues from context, target history, or a forecast prior that support expectations about future behavior. As shown in Figure~\ref{fig:castreasoner-motivation}, such evidence helps establish which changes are plausible, when they may occur, and why the available information supports them. Without this grounding, reasoning may miss key cues or introduce misleading information, impairing forecast generation or refinement at individual future intervals.

We argue that the fundamental challenge in LLM reasoning-driven context-aware time series forecasting is how to extract predictive evidence from contextual features and use it to guide forecasting decisions. To begin with, evidence extraction requires identifying contextual features relevant to the target and forecast horizon. Contextual influences vary across temporal scales and conditions~\cite{huang2025exogenous}, so reasoning must interpret cues against the target history to distinguish useful evidence from redundant or misleading information. Beyond this, evidence must inform the timing, direction, and magnitude of future changes. When a forecast prior is available, reasoning must determine which implications it already captures and where adjustments are justified. These challenges motivate us to design an evidence-grounded reasoning process connecting contextual interpretation with adaptive forecasting strategies and complete numerical predictions.

In this paper, we propose CastReasoner, an evidence-grounded reasoning model for context-aware time series forecasting. Its structured chain of thought (CoT) extracts predictive evidence from contextual features, infers the future dynamics suggested by this evidence, and assesses them against the forecast prior to decide whether to retain, revise, or reconstruct it. To enable this capability, we design a reasoning-centric two-stage post-training framework including SFT and RL. First, we synthesize reasoning trajectories through controllable multi-teacher guidance and select high-quality ones using a novel student-aware difficulty metric. Second, we design fine-grained multi-objective rewards to improve reasoning and forecasting. A convergence-guided data balancing strategy addresses heterogeneous learning needs across domains in both stages. Through this two-stage training, CastReasoner improves both reasoning quality and predictive accuracy.

We evaluate CastReasoner on eight real-world datasets across energy, urban mobility, air quality, retail sales, and industrial monitoring, with consistent gains over representative forecasting baselines. Ablation studies support the complementary roles of SFT and RL and the design of our reasoning process. Further analyses show that removing or disrupting reasoning traces increases forecasting errors, highlighting the importance of intermediate reasoning for prediction. CastReasoner also selects adaptive strategies according to the scope of predictive evidence. We hope these findings inspire further research into how evidence-grounded reasoning improves context-aware time series forecasting.

Our main contributions are summarized as follows:

\begingroup
\setlength{\listisep}{0pt}
\setlength{\partopsep}{0pt}
\begin{itemize}
\setlength{\itemsep}{0pt}
\setlength{\parsep}{0pt}
  \item We argue that effective context-aware forecasting requires interpreting predictive evidence relative to forecast priors. This motivates fine-grained reasoning that connects contextual evidence, inferred future dynamics, and adaptive forecasting decisions at individual future intervals.
  \item We propose CastReasoner, an evidence-grounded reasoning LLM that extracts predictive evidence from contextual features and forecast priors to guide adaptive forecasting strategies. We design a novel reasoning-centric two-stage post-training framework including SFT and RL.
  \item We evaluate CastReasoner on eight forecasting datasets and demonstrate consistent gains over representative baselines. Ablations and analyses of reasoning traces support the effectiveness of its training components and the role of intermediate reasoning in numerical prediction.
\end{itemize}
\endgroup

\begin{figure*}[!t]
  \centering
  \includegraphics[width=\textwidth]{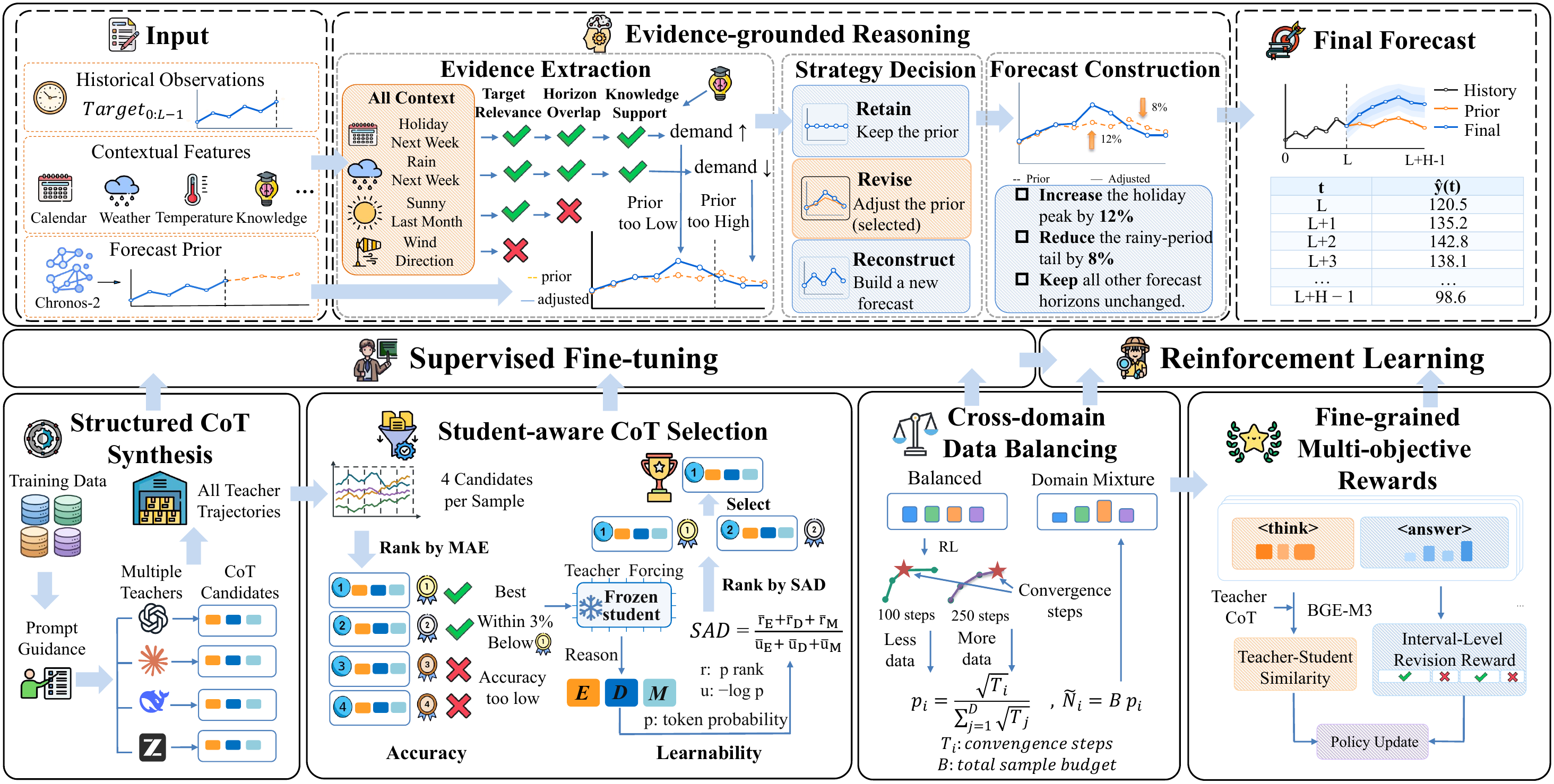}
  \caption{Overview of CastReasoner. Evidence-grounded forecasting links history, context, and a forecast prior to evidence extraction, strategy decision, and forecast construction. Structured CoT synthesis, student-aware CoT selection, cross-domain data balancing, and fine-grained multi-objective RL rewards form the post-training pipeline.}
  \Description{The upper panel connects target history, context, and a forecast prior to evidence analysis, selection among retain, revise, and reconstruct, and a complete forecast. The lower panel shows teacher-guided candidate trajectories selected by accuracy and learnability; an equal-mixture pilot estimates each domain's best-validation step, whose square root determines fixed quotas for SFT and RL; reinforcement learning combines fine-grained forecasting rewards with a group-relative policy update and KL regularization.}
  \label{fig:castreasoner-overview}
\end{figure*}

\section{Related Work}
We review how LLM-based time series forecasting incorporates and interprets context, followed by post-training approaches that develop reasoning for numerical prediction.

\subsection{Context-aware Time Series Forecasting \\ with Large Language Models}
Early LLM-based time series forecasting treated numerical histories as language sequences, enabling forecasts through text completion~\cite{xue2024promptcast,gruver2023llmtime}. Subsequent efforts sought closer alignment between temporal inputs and language representations~\cite{cao2024tempo}. Reprogramming mapped historical patterns into a pretrained language model's embedding space~\cite{jin2024timellm}. Symbolic discretization made numerical values compatible with contextual descriptions~\cite{tao2025tokencast}. Vector-injected demonstrations further adapted frozen LLMs to temporal inputs through in-context learning~\cite{zhang2026vectoricl}. As the focus broadened from representing histories to incorporating external influences, language knowledge helped model interactions between target series and exogenous variables~\cite{huang2025exogenous}. Retrieval extended the available context beyond a local history, drawing on related series and text for cross-city forecasting~\cite{jiang2026crosscity}. Geographic images offered a complementary source of spatial information for traffic prediction~\cite{chen2026visionst}. Beyond supplying auxiliary inputs, context also became an object of explicit analysis. Event-based forecasting interpreted external reports to identify influences on future values and used reflection to refine the resulting predictions~\cite{wang2024news}. Recent temporal reasoning further connects observed patterns and contextual features to expected future behavior through intermediate analysis~\cite{cheng2026timereasoner}. Together, these developments connect contextual interpretation and temporal reasoning with the generation of numerical forecasts.

\subsection{Reasoning-Oriented Post-Training}
Post-training for time series forecasting increasingly targets the reasoning process behind numerical predictions~\cite{cao2025conversational}. Supervised adaptation uses synthetic trajectories to demonstrate how temporal analysis precedes numerical prediction. Reinforcement learning then evaluates generated trajectories through forecasting-specific, multi-objective rewards~\cite{zhou2026timer1}. This extends broader progress in using verifiable outcomes to improve reasoning beyond imitation~\cite{shao2024deepseekmath,deepseek2025r1}. Yet outcome correctness does not guarantee that intermediate temporal judgments are sound. Process-verifiable annotations therefore introduce explicit checks of intermediate reasoning, with task- and difficulty-aware schedules tailoring learning to different temporal reasoning demands~\cite{zhou2026veritime}. For context-aware forecasting, the training objective also expands from producing future values to deciding when an existing forecast requires intervention. Revision trajectories and verifiable rewards teach models to preserve, revise, or disregard foundation-model priors according to multimodal context~\cite{liu2026posttime}. Beyond forecasting, table reasoning combines self-verified trajectories with reinforcement fine-tuning and uncertainty-based path selection to improve reasoning quality and stability~\cite{zhang2026star}. However, supervision for connecting contextual evidence with forecast priors and interval-specific forecasting decisions remains limited. To bridge this gap, CastReasoner develops evidence-grounded reasoning through demonstrations selected for accuracy and student learnability, followed by RL with fine-grained rewards.

\section{Methodology}
We introduce CastReasoner's evidence-grounded reasoning process and our reasoning-centric two-stage post-training framework with supervised fine-tuning (SFT) and reinforcement learning (RL).

\subsection{Problem Definition}
Let $\mathcal{D}=\{(X_i,\mathbf{y}_i)\}_{i=1}^{N}$ be a context-aware time series forecasting dataset, where each input is $X_i=(\mathbf{x}^{(i)}_{-L:-1},\mathcal{C}^{(i)},\mathbf{f}^{(0,i)})$. It contains the target history $\mathbf{x}^{(i)}_{-L:-1}$, features available for the forecast horizon $\mathcal{C}^{(i)}$, and a forecast prior $\mathbf{f}^{(0,i)}$; $\mathbf{y}_i\in\mathbb{R}^{H}$ is the future target over the next $H$ steps. CastReasoner serializes $X_i$ with a fixed prompt $P$, generates $T_i=\operatorname{LLM}_{\theta}(P(X_i))$, and extracts the complete numerical forecast as $\hat{\mathbf{y}}_i=g(T_i)$. The future target is reserved for offline trajectory selection, supervision, and reward evaluation.

\subsection{CastReasoner Overview}
Figure~\ref{fig:castreasoner-overview} shows how CastReasoner combines evidence-grounded forecasting with reasoning-centric post-training. It jointly examines history, context, and a forecast prior generated by Chronos-2~\cite{ansari2025chronos2} to identify evidence about future dynamics while excluding irrelevant contextual features. It then selects how much prior structure to preserve and constructs a complete forecast. SFT builds this capability from controllable teacher trajectories selected for forecast accuracy and student learnability. RL further improves reasoning and forecasting through group relative policy optimization (GRPO)~\cite{shao2024deepseekmath} with fine-grained multi-objective rewards.

\subsection{Evidence-Grounded Reasoning Process}
CastReasoner uses three reasoning stages: evidence analysis identifies predictive influences and their implications for future dynamics; strategy selection determines whether to retain, revise, or reconstruct the forecast prior; forecast construction turns the selected strategy and evidence into a complete numerical forecast.

During evidence analysis, CastReasoner extracts predictive evidence from exogenous variables, calendar signals, and domain information in $\mathcal{C}$. It assesses each condition against the target's meaning, observed history, and domain knowledge, retaining plausible influences within the forecast horizon and excluding irrelevant conditions. For example, high afternoon temperatures may affect electricity load through cooling demand, whereas events whose effects do not overlap the forecast horizon provide no relevant evidence. From the retained evidence, the model infers the direction, relative strength, onset, duration, and shape of future changes, preserving uncertainty where support is limited. It compares these dynamics with the corresponding prior intervals to identify effects that are already represented, understated, overstated, or mistimed, providing the basis for subsequent strategy selection.

During strategy selection, CastReasoner determines the extent of numerical intervention. It selects \texttt{retain} when the prior agrees with the evidence or support for correction is insufficient; \texttt{revise} when evidence warrants a bounded adjustment while useful prior structure remains; and \texttt{reconstruct} when the prior's overall shape conflicts with inferred dynamics, requiring a new forecast from history and context. This choice links the scope and strength of the evidence to the prior structure preserved in the prediction.

During forecast construction, the model specifies how the selected strategy and its supporting evidence produce all $H$ numerical values. It copies the prior for \texttt{retain}, specifies supported corrections and preserved behavior for \texttt{revise}, or explains how history and context support a replacement trajectory for \texttt{reconstruct}. The method connects inferred dynamics to the forecast level, shape, and timing, translating qualitative influences into numerical choices. Every strategy returns the complete forecast, including unchanged positions, indexed from $0$ to $H-1$ with precision appropriate to the target scale. The resulting trajectory $o=(d,\mathcal{E},m,\hat{\mathbf{y}})$ records the selected strategy $d$, evidence analysis $\mathcal{E}$, forecast method $m$, and complete numerical forecast $\hat{\mathbf{y}}\in\mathbb{R}^{H}$.

\subsection{Trajectory Distillation for Reasoning Capability Initialization}
\subsubsection{Structured CoT Synthesis}
To benefit from the different reasoning capabilities of multiple LLMs, we use four teachers to generate structured reasoning trajectories with alternative evidence interpretations and forecasting strategies for the same sample: DeepSeek-V4-Flash~\cite{deepseek2026v4}, GPT-6.1-Sol~\cite{openai2026gpt61sol}, GLM-5.3-Flash~\cite{zai2026glm53flash}, and Claude Opus 5.5~\cite{anthropic2026opus55}. Each generates one trajectory from the same input $X$ available to the student; future targets are used only for offline evaluation of the generated forecasts.
Prompt instructions prescribe the three-stage structure in Section~3.3 and require all $H$ forecast values: teachers first analyze evidence and future dynamics, then select a strategy, and finally explain forecast construction before giving the numerical answer. The shared structure keeps reasoning fields comparable while preserving diverse interpretations across teachers, facilitating subsequent student-aware trajectory selection.

\subsubsection{Student-aware CoT Selection}
We discard invalid candidates and rank the remaining forecasts by mean absolute error (MAE), $e_k=\operatorname{MAE}(\hat{\mathbf{y}}^{(k)},\mathbf{y})$. This student-aware criterion favors accurate trajectories that expose useful reasoning opportunities for the current student. We assess learnability with student-aware difficulty (SAD), favoring tokens with low student probabilities that rank near the vocabulary top: low probabilities leave room for learning, while high ranking keeps reasoning within the student's initial capability. For token $z_j^{(k)}$, let $p_{k,j}=p_{\theta_0}(z_j^{(k)}\mid X,z_{<j}^{(k)})$ and $r_{k,j}$ be its probability and one-based rank under the frozen student. Thus, the score reflects learnability from the current student's perspective. Within each field $s\in\mathcal{F}=\{d,\mathcal{E},m\}$, we average clipped ranks $\min\{r_{k,j},r_{\max}\}$ and the corresponding negative log probabilities $-\log p_{k,j}$ to obtain $\bar r_{k,s}$ and $\bar u_{k,s}$. SAD combines them as:
\begin{equation}
  \operatorname{SAD}_k=\frac{\sum_{s\in\mathcal{F}}\bar r_{k,s}}{\sum_{s\in\mathcal{F}}\bar u_{k,s}}.
\end{equation}

Field means prevent longer fields from dominating. Lower token probabilities yield larger negative log probabilities, increasing the denominator, while tokens ranked closer to the vocabulary top reduce the numerator. We score only visible reasoning tokens, excluding final forecast tokens. Among the two best-MAE candidates, $k_1$ and $k_2$, we select the lower-SAD candidate only if their strategies agree and $e_{k_2}\leq(1+\delta)e_{k_1}$; otherwise, we choose $k_1$. The selected trajectories supervise SFT; RL starts from the final SFT checkpoint.

\begin{table*}[t]
\centering
\caption{Overall forecasting performance of CastReasoner (9B) and baselines on eight datasets, measured by nMAE and nMSE. The best and second-best results in each column are shown in \textbf{bold} and \underline{underlined}, respectively.}
\label{tab:main-results}
\begingroup
\normalsize
\setlength{\tabcolsep}{1.2pt}
\renewcommand{\arraystretch}{1.0}
\begin{tabular}{@{}ccccccccccccccccc@{}}
\toprule
\multirow[c]{2}{*}{\textbf{Model}} & \multicolumn{2}{c}{\textbf{Load}} & \multicolumn{2}{c}{\textbf{Price}} & \multicolumn{2}{c}{\textbf{Solar}} & \multicolumn{2}{c}{\textbf{Wind}} & \multicolumn{2}{c}{\textbf{BS}} & \multicolumn{2}{c}{\textbf{BAQ}} & \multicolumn{2}{c}{\textbf{HS}} & \multicolumn{2}{c}{\textbf{MetroPT-3}} \\
\cmidrule(lr){2-3}\cmidrule(lr){4-5}\cmidrule(lr){6-7}\cmidrule(lr){8-9}\cmidrule(lr){10-11}\cmidrule(lr){12-13}\cmidrule(lr){14-15}\cmidrule(lr){16-17}
 & nMAE & nMSE & nMAE & nMSE & nMAE & nMSE & nMAE & nMSE & nMAE & nMSE & nMAE & nMSE & nMAE & nMSE & nMAE & nMSE \\
\midrule
ARIMA & 1.8470 & 7.5514 & 0.5963 & 1.0713 & 1.9798 & 5.8391 & 1.0234 & 1.7691 & 0.9491 & 1.7476 & 0.5808 & 0.9649 & 1.2498 & 3.6027 & 1.2446 & 2.4615 \\
Prophet & 0.6074 & 0.7953 & 0.4398 & 0.6305 & 1.3154 & 3.8053 & 1.3100 & 3.0755 & 0.4855 & 0.4245 & 0.3699 & 0.4783 & 0.9538 & 1.6255 & 0.4238 & 0.4370 \\
PatchTST & 0.3813 & 0.2884 & 0.3693 & 0.6078 & 0.3315 & 0.3192 & 0.8658 & 1.1368 & 0.2783 & 0.1760 & 0.6104 & 0.8065 & 0.9748 & 1.6576 & 0.9381 & 1.2385 \\
iTransformer & 0.4431 & 0.3644 & 0.4289 & 0.7556 & 0.3797 & 0.3688 & 0.8724 & 1.1412 & 0.3040 & 0.1971 & 0.6395 & 0.8407 & 0.9245 & 1.5275 & 0.9183 & 1.1653 \\
TimeXer & 0.3931 & 0.3022 & 0.4041 & 0.7116 & 0.4454 & 0.4239 & 0.8569 & 1.1071 & 0.2832 & 0.1815 & 0.6557 & 0.8716 & 0.9606 & 1.8368 & 0.8939 & 1.0758 \\
TiDE & 0.4281 & 0.3440 & 0.4144 & 0.7180 & 0.4098 & 0.4040 & 0.8832 & 1.1692 & 0.3187 & 0.2128 & 0.6753 & 0.9266 & 1.0566 & 1.8741 & 0.9444 & 1.2767 \\
Chronos-bolt & 0.3937 & 0.3271 & 0.3640 & 0.5970 & 0.2670 & 0.2895 & 0.8217 & 1.0543 & 0.2373 & 0.1467 & 0.4982 & 0.6467 & 0.9435 & 1.4298 & 0.6794 & 0.8232 \\
Chronos-2 & 0.2624 & 0.1743 & 0.3323 & 0.5921 & 0.2429 & 0.2451 & 0.7108 & 0.9143 & 0.2146 & 0.1213 & 0.1812 & 0.0785 & \underline{0.8007} & \underline{1.2508} & 0.2618 & 0.1573 \\
TimesFM-2.5 & 0.3805 & 0.3059 & 0.3812 & 0.6893 & 0.2911 & 0.3260 & 0.8646 & 1.2377 & 0.2473 & 0.1582 & 0.5423 & 0.7715 & 0.8986 & 1.6772 & 0.5747 & 0.6210 \\
Moirai & 0.5339 & 0.5504 & 0.3993 & 0.7540 & 0.6728 & 0.8307 & 0.9497 & 1.4597 & 0.5169 & 0.6415 & 0.4959 & 0.6081 & 0.8715 & 1.3929 & 0.7621 & 0.9008 \\
Sundial & 0.3712 & 0.2885 & 0.3947 & 0.7026 & 0.2895 & 0.3015 & 0.8435 & 1.0766 & 0.2488 & 0.1591 & 0.5514 & 0.7036 & 0.9113 & 1.3263 & 0.5627 & 0.5763 \\
TimeReasoner & 0.2836 & 0.1885 & 0.2895 & \underline{0.3996} & 0.1765 & 0.1548 & 0.5366 & 0.5754 & \underline{0.1924} & \underline{0.1025} & 0.1395 & \underline{0.0488} & 0.8283 & 1.3929 & 0.2223 & \underline{0.1349} \\
Time-R1 & 0.2367 & 0.1305 & \underline{0.2829} & 0.4455 & 0.1732 & 0.1316 & 0.4713 & 0.4637 & 0.1971 & 0.1055 & 0.1400 & 0.0523 & 0.8043 & 1.2673 & 0.2273 & 0.1376 \\
PostTime & \underline{0.2318} & \underline{0.1283} & 0.2851 & 0.4532 & \underline{0.1703} & \underline{0.1298} & \underline{0.4647} & \underline{0.4521} & 0.2014 & 0.1081 & \underline{0.1375} & 0.0514 & 0.8106 & 1.2816 & \underline{0.2154} & 0.1427 \\
\midrule
\textbf{CastReasoner} & \textbf{0.2123} & \textbf{0.1101} & \textbf{0.2623} & \textbf{0.3658} & \textbf{0.1579} & \textbf{0.1277} & \textbf{0.4443} & \textbf{0.3821} & \textbf{0.1773} & \textbf{0.0858} & \textbf{0.1290} & \textbf{0.0478} & \textbf{0.7215} & \textbf{1.0828} & \textbf{0.1877} & \textbf{0.0998} \\
\bottomrule
\end{tabular}
\endgroup
\end{table*}

\subsection{Fine-Grained Multi-Objective Rewards \\for Reasoning Capability Enhancement}
\subsubsection{Cross-Domain Data Balancing}
To address heterogeneous learning needs across domains~\cite{liu2024unitime}, we design a convergence-guided balancing strategy. We first train a pilot model on an equally sampled mixture of all $D$ domains. Let $L_i(t)=\alpha_{\mathrm{val}}\operatorname{nMAE}_i(t)+(1-\alpha_{\mathrm{val}})\operatorname{nMSE}_i(t)$ denote domain $i$'s validation loss at evaluation step $t$, and define its convergence step $T_i$ as the step at which $L_i(t)$ reaches its minimum during the pilot. We use $T_i$ to determine the sampling probability and target quota for each domain under a fixed total training sample budget $B$:
\begin{equation}
  p_i=\frac{\sqrt{T_i}}{\sum_{j=1}^{D}\sqrt{T_j}},
  \qquad \widetilde{N}_i=Bp_i.
\end{equation}

Since $p_i$ increases with $T_i$, slower-converging domains receive larger quotas. Square-root scaling reduces convergence disparities while keeping quota gaps moderate, leading to better training performance. The quotas are fixed before formal training and remain unchanged throughout SFT and RL.

\subsubsection{Fine-grained Multi-objective Rewards}
We combine five rewards for output validity, forecast accuracy, baseline improvement, localized prior revision, and learning from teacher reasoning.

\paragraph{Format Reward.}
We penalize responses that violate the required reasoning structure or numerical forecast format:
\begin{equation}
  R_{\mathrm{format}}(o)=-\lambda_f\bigl(1-V(o)\bigr),
\end{equation}
where $V(o)$ indicates whether the \texttt{<think>} block follows the three stages and the \texttt{<answer>} provides all $H$ finite predictions; $\lambda_f$ sets the penalty. It is zero for valid outputs and negative otherwise; all other rewards evaluate only valid responses.

\paragraph{Forecast Accuracy Reward.}
The accuracy reward is the negative clipped normalized forecasting loss:
\begin{equation}
  R_{\mathrm{accuracy}}=-L_c(\hat{\mathbf{y}}),
\end{equation}
where $L_c(\mathbf{f})$ is the clipped loss for forecast $\mathbf{f}$. The scale is $s_y=\max\{\operatorname{Std}(\mathbf{y}),\rho H^{-1}\sum_t|y_t|,\epsilon_s\}$. We define normalized absolute and squared errors as $a(\mathbf{f})=H^{-1}\sum_t|f_t-y_t|/s_y$ and $b(\mathbf{f})=H^{-1}\sum_t(f_t-y_t)^2/s_y^2$. Combining them as $L(\mathbf{f})=\alpha a(\mathbf{f})+(1-\alpha)b(\mathbf{f})$ and clipping each component at $C$ gives $L_c(\mathbf{f})$.

\paragraph{Simple Baseline Outperformance Reward.}
We reward improvement over simple forecasting baselines using:
\begin{equation}
  R_{\mathrm{uplift}}=\lambda_U\min\left\{
    \left[\frac{g_u-\tau_g}{c_g}\right]_0^1,
    \left[\frac{\Delta_u-\tau_\Delta}{c_\Delta}\right]_0^1\right\},
\end{equation}
where $\Delta_u=L_{\mathrm{ref}}-L_c(\hat{\mathbf{y}})$ and $g_u=\Delta_u/\max\{L_{\mathrm{ref}},\epsilon_U\}$ measure absolute and relative improvement. Here $L_{\mathrm{ref}}$ is the lowest clipped loss among last-value, recent-mean, seasonal-copy, and Ridge regression~\cite{hoerl1970ridge} baselines, built without the evaluated future target. The operator $[v]_a^b=\min\{b,\max\{a,v\}\}$ clips its input; $\lambda_U$ scales the reward, $\tau_g,\tau_\Delta$ are thresholds, and $c_g,c_\Delta$ are ranges. Both gains must exceed their thresholds for a positive reward.

\paragraph{Interval-Level Prior Revision Reward.}
We reward beneficial local forecast corrections within evidence-affected intervals:
\begin{equation}
  R_{\mathrm{interval}}=\lambda_{\mathrm{loc}}\sum_{k=1}^{K}w_k[\Delta_k]_{-c_{\mathrm{loc}}}^{c_{\mathrm{loc}}}-\lambda_O D_O,
\end{equation}
where $\mathcal{I}=\{I_1,\ldots,I_K\}$ is the set of target intervals. For position set $A$, $L_A$ is the loss $L$ averaged over its positions. The local improvement is $\Delta_k=L_{I_k}(\mathbf{f}^{(0)})-L_{I_k}(\hat{\mathbf{y}})$, weighted by $w_k=|I_k|/\sum_j|I_j|$. The complement $O$ covers positions outside the intervals; $D_O$ is the positive increase in $L_O$, clipped at $c_O$ and zero if $O$ is empty. The parameters $\lambda_{\mathrm{loc}},c_{\mathrm{loc}},\lambda_O$ set the local scale, clipping bound, and outside weight. DeepSeek-V4-Flash~\cite{deepseek2026v4} annotates candidate intervals from context without using ground-truth future targets as input. For training samples without annotated target intervals, this reward term is set to zero during policy optimization.

\paragraph{Teacher-Guided Reasoning Preference Reward.}
We guide reasoning toward effective teacher demonstrations:
\begin{equation}
  R_{\mathrm{reasoning}}=\lambda_T Q\left[\frac{s_{\mathrm{CoT}}-\tau_T}{1-\tau_T}\right]_0^1,
\end{equation}
where $s_{\mathrm{CoT}}$ is the maximum cosine similarity between student and teacher chains of thought (CoTs), computed by a frozen BGE-M3 encoder~\cite{chen2024m3} using teacher CoTs whose forecasts improve upon the prior in MAE. The indicator $Q$ requires valid CoTs and student loss $L(\hat{\mathbf{y}})\leq L(\mathbf{f}^{(0)})$; $\lambda_T$ and $\tau_T$ set the scale and threshold. Combining all reward components, the overall reward is:
\begin{equation}
  R(o)=R_{\mathrm{format}}(o)+V(o)\bigl(R_{\mathrm{accuracy}}+R_{\mathrm{uplift}}+R_{\mathrm{interval}}+R_{\mathrm{reasoning}}\bigr).
\end{equation}

These signals jointly guide valid, accurate forecasts and evidence-grounded reasoning during RL, combining overall predictive quality with localized revision and teacher guidance.

\section{Experiments}
We evaluate forecasting accuracy across eight datasets, examine the reasoning and training components, and analyze how contextual evidence shapes prediction decisions.

\subsection{Experimental Setup}
\paragraph{Datasets.}
For long-term forecasting, we use four energy datasets, Load, Price, Solar, and Wind, from the Global Energy Forecasting Competition 2014 (GEFCom2014)~\cite{hong2016gefcom}. For short-term forecasting, we use four datasets catalogued by the UCI Machine Learning Repository: Bike Sharing (BS)~\cite{fanaee2014bike}, Beijing Multi-Site Air Quality (BAQ)~\cite{zhang2017baq}, Hierarchical Sales Data (HS)~\cite{mancuso2021hs}, and the MetroPT-3 railway compressor dataset~\cite{davari2021metropt}, spanning the Urban Mobility, Environment, Retail, and Industrial Monitoring domains, respectively. We report nMAE and nMSE; Appendix~\ref{app:datasets} describes dataset periods, inputs, forecasting windows, and metric definitions.

\paragraph{Baselines.}
We compare four baseline categories. Statistical methods are ARIMA and Prophet~\cite{box1970timeseries,taylor2018prophet}. Deep learning methods include PatchTST, iTransformer, TimeXer, and TiDE~\cite{nie2023patchtst,liu2024itransformer,wang2024timexer,das2023tide}. Foundation models include Chronos-bolt, Chronos-2, TimesFM-2.5, Moirai, and Sundial~\cite{ansari2024chronosbolt,ansari2024chronos,das2024timesfm,woo2024moirai,liu2025sundial}. In particular, Chronos-2 is evaluated with the same available numerical covariates used by CastReasoner to ensure a fair comparison. Reasoning LLMs include TimeReasoner, Time-R1, and PostTime~\cite{cheng2026timereasoner,zhou2026timer1,liu2026posttime}.

\paragraph{Implementation Details.}
We use Qwen3.5-9B~\cite{qwen2026qwen35} for main results and Qwen3.5-4B for auxiliary experiments. SFT uses 2,000 trajectories for three epochs at learning rate $2\times10^{-5}$. GRPO~\cite{shao2024deepseekmath} RL uses two epochs at learning rate $10^{-6}$, $\epsilon=0.2$, and $\beta=0.001$. Training uses VeRL~\cite{sheng2024hybridflow} and vLLM~\cite{kwon2023vllm} on four 80GB H100 GPUs. During training, samples from all eight datasets are mixed and randomly shuffled. Detailed hyperparameters for both training stages and the reward functions are listed in Appendix~\ref{app:experimental-parameters}.

\begin{table}[t]
\centering
\caption{Ablations of CastReasoner (4B) on Load, Wind, and BAQ. We remove evidence analysis, strategy selection, forecast construction, SFT, or RL one at a time.}
\label{tab:ablation}
\begingroup
\normalsize
\setlength{\tabcolsep}{1.2pt}
\renewcommand{\arraystretch}{1.08}
\begin{tabular}{@{}crrrrrr@{}}
\toprule
\multirow[c]{2}{*}{\textbf{Method}} & \multicolumn{2}{c}{\textbf{Load}} & \multicolumn{2}{c}{\textbf{Wind}} & \multicolumn{2}{c}{\textbf{BAQ}} \\
\cmidrule(lr){2-3}\cmidrule(lr){4-5}\cmidrule(lr){6-7}
 & nMAE & nMSE & nMAE & nMSE & nMAE & nMSE \\
\midrule
CastReasoner & \textbf{0.2120} & \textbf{0.1299} & \textbf{0.4394} & \textbf{0.3978} & \textbf{0.1317} & \textbf{0.0506} \\
\midrule
w/o Analysis & 0.2504 & 0.1670 & 0.4635 & 0.4266 & 0.1456 & 0.0667 \\
w/o Decision & 0.2473 & 0.1616 & 0.4481 & 0.4037 & 0.1353 & 0.0587 \\
w/o Construction & 0.2469 & 0.1632 & 0.4539 & 0.4078 & 0.1394 & 0.0573 \\
\midrule
w/o SFT & 0.2625 & 0.1742 & 0.6720 & 0.8196 & 0.1802 & 0.0782 \\
w/o RL & 0.2485 & 0.1424 & 0.4634 & 0.4254 & 0.1503 & 0.0537 \\
\bottomrule
\end{tabular}
\endgroup
\end{table}

\begin{table}[t]
\centering
\caption{Reasoning-trace interventions on Load, Wind, and BAQ. With forecasting inputs fixed, we compare original traces against removed, shuffled, and corrupted traces.}
\label{tab:reasoning-intervention}
\begingroup
\normalsize
\setlength{\tabcolsep}{1.2pt}
\renewcommand{\arraystretch}{1.08}
\begin{tabular}{@{}ccccccc@{}}
\toprule
\multirow[c]{2}{*}{\textbf{Method}} & \multicolumn{2}{c}{\textbf{Load}} & \multicolumn{2}{c}{\textbf{Wind}} & \multicolumn{2}{c}{\textbf{BAQ}} \\
\cmidrule(lr){2-3}\cmidrule(lr){4-5}\cmidrule(lr){6-7}
 & nMAE & nMSE & nMAE & nMSE & nMAE & nMSE \\
\midrule
Original & \textbf{0.2120} & \textbf{0.1299} & \textbf{0.4394} & \textbf{0.3978} & \textbf{0.1317} & \textbf{0.0506} \\
No Think & 0.2331 & 0.1360 & 0.4526 & 0.4087 & 0.1431 & 0.0611 \\
Shuffled Think & 0.2432 & 0.1436 & 0.4677 & 0.4213 & 0.1589 & 0.0751 \\
Corrupted Think & 0.2496 & 0.1514 & 0.4739 & 0.4332 & 0.1657 & 0.0824 \\
\bottomrule
\end{tabular}
\endgroup
\end{table}

\subsection{Main Results}
Table~\ref{tab:main-results} reports nMAE and nMSE on eight datasets. CastReasoner ranks first on both metrics across all eight datasets. Foundation models extrapolate learned patterns; reasoning baselines interpret context. PostTime achieves eight second-best results, yet CastReasoner improves on it across all sixteen metrics. On MetroPT-3, its 30.1\% nMSE reduction over PostTime suggests better control of large errors, while the narrow Solar nMSE gap shows that benefits vary across domains. These comparisons support linking fine-grained predictive evidence to the decision to retain, revise, or reconstruct the prior. CastReasoner outperforms all baselines on datasets with strong periodicity, such as Load, and those with weaker periodicity, such as Wind. On Wind, it reduces nMAE and nMSE by 37.5\% and 58.2\% relative to Chronos-2 prior forecasts. This highlights its ability to extract predictive evidence from contextual features and relate it to future dynamics when recurring patterns alone provide limited guidance. It also performs strongly on datasets influenced by natural factors, such as Solar, and social factors, such as BS, where weather conditions and human activity shape future dynamics in different ways. Together, these results demonstrate CastReasoner's strong evidence-grounded forecasting capability across diverse temporal patterns and sources of contextual influence.

\subsection{Ablation Study}
Table~\ref{tab:ablation} compares reasoning-step and training-stage ablations of Qwen3.5-4B on Load, Wind, and BAQ to assess each component's contribution to forecasting accuracy.

\begin{figure}[t]
\centering
\includegraphics[width=\columnwidth]{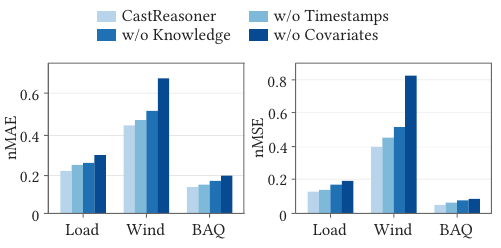}
\caption{Contextual-input ablations on Load, Wind, and BAQ. We separately remove timestamps, exogenous covariates, and domain knowledge to measure each source's contribution.}
\label{fig:context-ablation}
\end{figure}

\begin{figure}[t]
\centering
\includegraphics[width=\columnwidth]{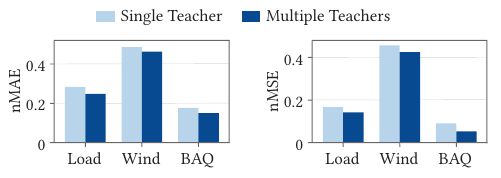}
\caption{Single- and multi-teacher SFT on Load, Wind, and BAQ. We compare one teacher with diverse teachers under the same trajectory budget and trajectory selection strategy.}
\label{fig:teacher-comparison}
\end{figure}

\subsubsection{Reasoning-Step Ablation}
To assess the three reasoning steps, we retrain the model after removing one step at a time. All domains worsen most without evidence analysis, showing that identifying relevant context and inferring its future effects is critical to forecasting. Strategy selection makes the model output an explicit decision---retain, revise, or reconstruct---and removing it also degrades performance. Forecast construction converts the decision into a complete horizon prediction; removing this step weakens the rationale-output link. Together, these results show that all three steps are essential to high-quality reasoning and forecasting; none can be omitted without compromising performance.

\subsubsection{Training-Stage Ablation}
Removing either stage hurts all domains, but omitting SFT causes the larger decline, especially on weakly periodic Wind. Teacher traces train the model to extract evidence, choose a prior-handling strategy, and produce a valid full-horizon forecast. Without SFT, the model must discover both its reasoning process and output format through outcome rewards, making exploration difficult. RL refines this SFT-taught policy using accuracy, uplift, interval, and teacher-guided rewards. Without RL, the structured process remains but receives less feedback about predictive accuracy. SFT establishes how to reason; RL improves how well the model applies that process to forecasting.

\subsection{In-Depth Analysis}
We analyze contextual inputs, reasoning quality, and adaptive strategy selection, then examine how SFT supervision and RL design shape evidence-grounded forecasting.

\begin{table}[t]
\centering
\caption{SFT trajectory selection: Random samples one of four teacher traces; Accuracy Only selects the lowest-MAE trace; Ours combines accuracy with student learnability.}
\label{tab:trajectory-selection}
\begingroup
\normalsize
\setlength{\tabcolsep}{1.2pt}
\renewcommand{\arraystretch}{1.08}
\begin{tabular}{@{}crrrrrr@{}}
\toprule
\multirow[c]{2}{*}{\textbf{Method}} & \multicolumn{2}{c}{\textbf{Load}} & \multicolumn{2}{c}{\textbf{Wind}} & \multicolumn{2}{c}{\textbf{BAQ}} \\
\cmidrule(lr){2-3}\cmidrule(lr){4-5}\cmidrule(lr){6-7}
 & nMAE & nMSE & nMAE & nMSE & nMAE & nMSE \\
\midrule
Random & 0.2604 & 0.1657 & 0.5105 & 0.5068 & 0.1839 & 0.0697 \\
Accuracy Only & 0.2528 & 0.1606 & 0.4714 & 0.4382 & 0.1586 & 0.0602 \\
Ours & \textbf{0.2485} & \textbf{0.1424} & \textbf{0.4634} & \textbf{0.4254} & \textbf{0.1503} & \textbf{0.0537} \\
\bottomrule
\end{tabular}
\endgroup
\end{table}

\begin{figure}[t]
\centering
\includegraphics[width=\columnwidth]{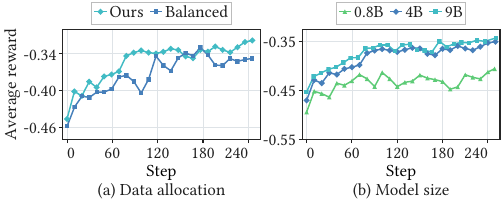}
\caption{Eight-domain mean RL reward: (a) equal (Balanced) versus convergence-guided (Ours) sampling; (b) 0.8B, 4B, and 9B students with fixed data allocation.}
\label{fig:reward-training}
\label{fig:cross-domain-reward}
\label{fig:model-size-reward}
\end{figure}

\subsubsection{Contextual Features Analysis}
Figure~\ref{fig:context-ablation} compares ablations of timestamps, covariates, and domain knowledge. Covariates matter most, especially on Wind, where their removal more than doubles nMSE. Weather signals describe changing conditions that historical repetition alone cannot capture. Domain knowledge helps interpret such signals, while timestamps place them within daily or seasonal patterns. Removing timestamps has a generally modest effect, suggesting that temporal cues complement but cannot replace external context. These findings support combining contextual sources to infer future dynamics and guide evidence-grounded forecasts across diverse temporal patterns and domains.

\subsubsection{Inference-Time Reasoning-Trace Intervention}
Table~\ref{tab:reasoning-intervention} changes the intermediate trace while fixing forecasting inputs. Across all domains, original traces perform best, followed by removed, shuffled, and corrupted traces. Removing reasoning deprives the model of useful guidance, while inconsistent or misleading traces can steer predictions toward unsupported changes. Their greater harm suggests that the model uses its intermediate reasoning when constructing forecasts. Together, these results show that coherent reasoning helps translate contextual evidence into well-supported forecast decisions, establishing a connection between intermediate inferences and the quality of final numerical forecasts.

\subsubsection{SFT Analysis}
Figure~\ref{fig:teacher-comparison} compares single- and multi-teacher supervision with the same trajectory budget and selection strategy. Multiple teachers improve both metrics across all three domains, with the largest relative gains on BAQ, providing diverse interpretations of contextual evidence. Table~\ref{tab:trajectory-selection} shows that accuracy-only selection improves over random selection, while incorporating student learnability performs best. Accurate forecasts can still be difficult to imitate; the joint criterion favors useful trajectories within the student's reach. Together, diverse teachers and joint selection improve supervision for evidence-grounded forecasting.

\subsubsection{RL Analysis}
Figure~\ref{fig:cross-domain-reward}(a) shows that convergence-guided quotas achieve higher reward through most of training than equal sampling under the same budget. Square-root allocation gives slower-converging domains more examples while limiting quota disparities. Table~\ref{tab:reward-ablation} shows that accuracy reward matters most, grounding optimization in forecast errors. Baseline outperformance encourages gains over elementary forecasts; interval feedback guides localized corrections; teacher preference reinforces effective reasoning. Format reward matters least, consistent with SFT establishing output structure. These results support convergence-guided allocation and complementary rewards for accurate, context-sensitive forecasting.

\subsubsection{Strategy Adaptation to Evidence Scope}
\label{sec:strategy-adaptation}
\begin{figure*}[t]
\begin{minipage}{\columnwidth}
\centering
\includegraphics[width=\columnwidth]{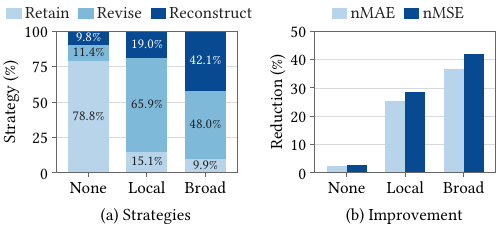}
\caption{By evidence group (None, Local, Broad): (a) student strategies; (b) relative nMAE/nMSE reduction over the prior.}
\label{fig:strategy-adaptation}
\end{minipage}\hfill
\begin{minipage}{\columnwidth}
\captionsetup{type=figure}
\centering
\includegraphics[width=\columnwidth]{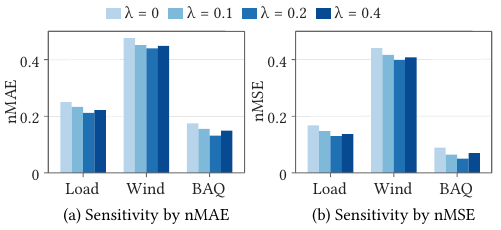}
\caption{Local interval-reward weight $\lambda$ sensitivity on Load, Wind, and BAQ, measured by nMAE and nMSE.}
\label{fig:interval-weight}
\end{minipage}
\end{figure*}

Four teacher LLMs independently label test samples by predictive evidence: none, localized impact, or broad impact affecting over 80\% of the forecast horizon; the most-voted label defines each group. CastReasoner receives no labels during inference. Figure~\ref{fig:strategy-adaptation} reports strategy distributions and relative nMAE/nMSE reductions over the prior using the main evaluation protocol. Without evidence, Retain dominates at 78.8\% with modest gains, showing restraint against unsupported changes. Localized evidence favors Revise (65.9\%) and substantially greater improvements. For broad impact, Reconstruct rises from 19.0\% to 42.1\%, alongside further error reductions. These results demonstrate adaptive strategy selection: preserving the prior without supporting evidence, revising affected intervals, and increasingly reconstructing forecasts when evidence has broad effects.

\begin{table}[t]
\centering
\caption{Reward ablation of CastReasoner (4B) on Load, Wind, and BAQ in nMAE/nMSE. Each variant removes one reward during RL; the full model uses all five.}
\label{tab:reward-ablation}
\begingroup
\normalsize
\setlength{\tabcolsep}{1.2pt}
\renewcommand{\arraystretch}{1.08}
\begin{tabular}{@{}crrrrrr@{}}
\toprule
\multirow[c]{2}{*}{\textbf{Method}} & \multicolumn{2}{c}{\textbf{Load}} & \multicolumn{2}{c}{\textbf{Wind}} & \multicolumn{2}{c}{\textbf{BAQ}} \\
\cmidrule(lr){2-3}\cmidrule(lr){4-5}\cmidrule(lr){6-7}
 & nMAE & nMSE & nMAE & nMSE & nMAE & nMSE \\
\midrule
CastReasoner & \textbf{0.2120} & \textbf{0.1299} & \textbf{0.4394} & \textbf{0.3978} & \textbf{0.1317} & \textbf{0.0506} \\
w/o $R_{\mathrm{format}}$ & 0.2135 & 0.1327 & 0.4413 & 0.4119 & 0.1396 & 0.0587 \\
w/o $R_{\mathrm{accuracy}}$ & 0.3786 & 0.2913 & 0.7145 & 0.6539 & 0.2431 & 0.1022 \\
w/o $R_{\mathrm{uplift}}$ & 0.2321 & 0.1375 & 0.4705 & 0.4277 & 0.1564 & 0.0748 \\
w/o $R_{\mathrm{interval}}$ & 0.2301 & 0.1329 & 0.4609 & 0.4211 & 0.1485 & 0.0721 \\
w/o $R_{\mathrm{reasoning}}$ & 0.2219 & 0.1337 & 0.4552 & 0.4168 & 0.1401 & 0.0597 \\
\bottomrule
\end{tabular}
\endgroup
\end{table}

\begin{figure*}[!t]
  \centering
  \includegraphics[width=\textwidth]{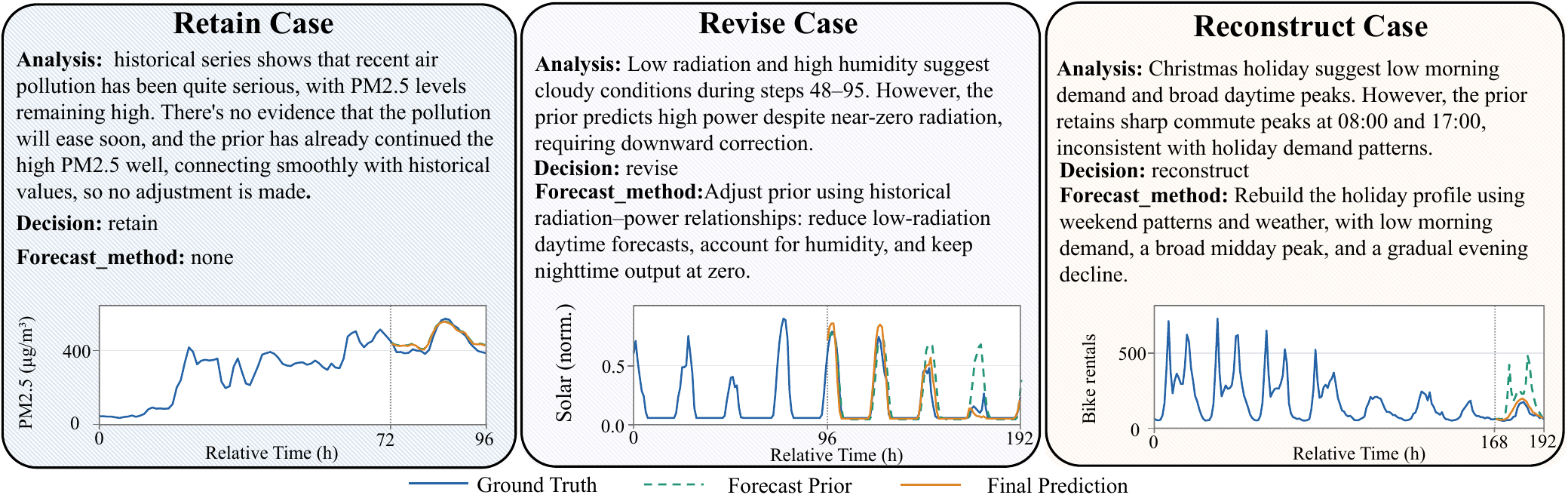}
  \caption{Reasoning and forecasts for BAQ (retain), Solar (revise), and BS (reconstruct). Curves compare ground truth, forecast priors, and final predictions, showing how contextual evidence guides the scope of forecast intervention.}
  \label{fig:case-study}
\end{figure*}

\subsection{Sensitivity Analysis}
We examine how model capacity and local interval-reward weight affect forecasting accuracy across domains.

\subsubsection{Model Size}
Figure~\ref{fig:model-size-reward}(b) compares reward trajectories for 0.8B, 4B, and 9B students. Both larger models sustain higher reward, suggesting that capacity helps learn evidence-grounded forecasting. The 0.8B curve recovers after a late dip, showing uneven gains from further training. Appendix~\ref{app:model-size} confirms lower errors for both larger students across all domains. Gains from 4B to 9B are selective: 9B improves both metrics on Price, BAQ, and HS, while 4B remains stronger on Solar and BS. Scaling thus improves overall forecasting, while gains depend on the dataset and evaluation metric.

\subsubsection{Local Interval Reward Weight}
Figure~\ref{fig:interval-weight} varies the interval-reward weight while retaining the other rewards. Both errors decrease across all domains as $\lambda$ rises from zero to 0.2, then increase at 0.4. The initial gains support directing feedback toward intervals where evidence warrants revising the prior. Global accuracy alone gives less explicit guidance about where to revise. The reversal at higher weight suggests that local revisions must remain balanced with complete-forecast accuracy. Moderate local feedback thus helps translate evidence into targeted corrections, with $\lambda=0.2$ giving the best joint accuracy among the tested settings.

\subsection{Case Study}
Figure~\ref{fig:case-study} illustrates how contextual evidence guides three prior-handling strategies. In BAQ, recent pollution remains high and no evidence suggests relief, supporting retention of the prior's level and trajectory. In Solar, low radiation and high humidity imply lower generation than the prior predicts, motivating downward revision within the affected forecast interval. In BS, holiday context contradicts the prior's sharp commute peaks, motivating reconstruction from weekend patterns and weather, with lower morning demand and a broader daytime peak. The cases link the extent of forecast intervention to the prior's agreement with the future dynamics inferred from contextual evidence. This reasoning guides both the direction and temporal scope of intervention.

\section{Conclusion}
CastReasoner links contextual evidence to forecast priors through evidence analysis, strategy selection, and forecast construction, guiding whether to retain, revise, or reconstruct forecasts. Its two-stage training combines multi-teacher demonstrations selected for accuracy and student learnability with fine-grained RL rewards and convergence-guided cross-domain data balancing. Experiments on eight datasets show consistent gains in nMAE and nMSE across diverse temporal patterns and contextual influences. Ablations highlight the complementary roles of the three reasoning steps and the two training stages. Reasoning-trace interventions further show that coherent intermediate reasoning matters for numerical prediction, while case studies illustrate how evidence guides the direction and scope of forecast intervention. We hope these findings advance evidence-grounded reasoning for context-aware time series forecasting and encourage further studies across diverse domains.

\bibliographystyle{ACM-Reference-Format}
\bibliography{references}

\appendix

\section{Experimental Setup}
\label{app:experimental-setup}

\subsection{Datasets}
\label{app:datasets}
\begin{table*}[t]
\centering
\caption{Dataset domains, sampling frequencies, time points, chronological split periods, and available exogenous-variable counts.}
\label{tab:dataset-overview}
\begingroup
\setlength{\tabcolsep}{1.8pt}
\renewcommand{\arraystretch}{1.05}
\begin{tabular}{@{}cccccccc@{}}
\toprule
\textbf{Dataset} & \textbf{Domain} & \textbf{Freq.} & \textbf{Time points} & \textbf{Train} & \textbf{Validation} & \textbf{Test} & \textbf{Exogenous Variables} \\
\midrule
Load & Energy & 1 h & 61,344 & 2005-01--2009-11 & 2009-11--2010-12 & 2010-12--2012-01 & 25 \\
Price & Energy & 1 h & 25,968 & 2011-01--2013-01 & 2013-01--2013-07 & 2013-07--2013-12 & 2 \\
Solar & Energy & 1 h & 19,704 & 2012-04--2013-10 & 2013-10--2014-02 & 2014-02--2014-07 & 12 \\
Wind & Energy & 1 h & 17,544 & 2012-01--2013-05 & 2013-05--2013-09 & 2013-09--2014-01 & 4 \\
BS & Urban Mobility & 1 h & 17,455 & 2011-01--2012-07 & 2012-07--2012-09 & 2012-09--2012-12 & 8 \\
BAQ & Environment & 1 h & 35,064 & 2013-03--2015-12 & 2015-12--2016-07 & 2016-07--2017-02 & 11 \\
HS & Retail & 1 day & 1,798 & 2014-01--2017-06 & 2017-06--2018-03 & 2018-03--2018-12 & 1 \\
MetroPT-3 & Industrial Monitoring & 1 min & 306,960 & 2020-02--2020-06 & 2020-06--2020-07 & 2020-07--2020-09 & 7 \\
\bottomrule
\end{tabular}
\endgroup
\end{table*}

\begin{table*}[!t]
\centering
\caption{Forecasting nMAE and nMSE of CastReasoner with 0.8B, 4B, and 9B student models across eight datasets. The best result for each metric is shown in bold.}
\label{tab:model-size}
\begingroup
\normalsize
\setlength{\tabcolsep}{1.2pt}
\renewcommand{\arraystretch}{1.0}
\begin{tabular*}{\textwidth}{@{\extracolsep{\fill}}c*{16}{c}@{}}
\toprule
\multirow[c]{2}{*}{\textbf{Model}} & \multicolumn{2}{c}{\textbf{Load}} & \multicolumn{2}{c}{\textbf{Price}} & \multicolumn{2}{c}{\textbf{Solar}} & \multicolumn{2}{c}{\textbf{Wind}} & \multicolumn{2}{c}{\textbf{BS}} & \multicolumn{2}{c}{\textbf{BAQ}} & \multicolumn{2}{c}{\textbf{HS}} & \multicolumn{2}{c}{\textbf{MetroPT-3}} \\
\cmidrule(lr){2-3}\cmidrule(lr){4-5}\cmidrule(lr){6-7}\cmidrule(lr){8-9}\cmidrule(lr){10-11}\cmidrule(lr){12-13}\cmidrule(lr){14-15}\cmidrule(lr){16-17}
 & nMAE & nMSE & nMAE & nMSE & nMAE & nMSE & nMAE & nMSE & nMAE & nMSE & nMAE & nMSE & nMAE & nMSE & nMAE & nMSE \\
\midrule
0.8B & 0.2769 & 0.1914 & 0.3352 & 0.5993 & 0.1743 & 0.1576 & 0.5057 & 0.5392 & 0.2083 & 0.1203 & 0.1472 & 0.0624 & 0.7848 & 1.2904 & 0.2445 & 0.1764 \\
4B & \textbf{0.2120} & 0.1299 & 0.2811 & 0.4066 & \textbf{0.1548} & \textbf{0.1226} & \textbf{0.4394} & 0.3978 & \textbf{0.1754} & \textbf{0.0822} & 0.1317 & 0.0506 & 0.7496 & 1.1267 & 0.1912 & \textbf{0.0944} \\
9B & 0.2123 & \textbf{0.1101} & \textbf{0.2623} & \textbf{0.3658} & 0.1579 & 0.1277 & 0.4443 & \textbf{0.3821} & 0.1773 & 0.0858 & \textbf{0.1290} & \textbf{0.0478} & \textbf{0.7215} & \textbf{1.0828} & \textbf{0.1877} & 0.0998 \\
\bottomrule
\end{tabular*}
\endgroup
\end{table*}

Table~\ref{tab:dataset-overview} summarizes the eight datasets in main-results order and gives time ranges at month resolution. Splits are strictly ordered and disjoint, with windows constructed independently; exogenous-variable counts include task-provided calendar or event inputs where applicable.

History/prediction/stride lengths (in sampling steps) are 96/96/24 for Load, Price, Solar, and Wind; 72/24/24 for BAQ; 120/30/30 for MetroPT-3; 30/7/7 for HS; and 168/24/24 for BS.

The exogenous variables are: Load, \textnormal{w1--w25}; Price, \textnormal{forecasted\_\allowbreak total\_\allowbreak load} and \textnormal{forecasted\_\allowbreak zonal\_\allowbreak load}; Solar, \textnormal{VAR78}, \textnormal{VAR79}, \textnormal{VAR134}, \textnormal{VAR157}, \textnormal{VAR164}, \textnormal{VAR165}, \textnormal{VAR166}, \textnormal{VAR167}, \textnormal{VAR169}, \textnormal{VAR175}, \textnormal{VAR178}, and \textnormal{VAR228}; Wind, \textnormal{U10}, \textnormal{V10}, \textnormal{U100}, and \textnormal{V100}; BS, \textnormal{holiday}, \textnormal{weekday}, \textnormal{workingday}, \textnormal{weathersit}, \textnormal{temp}, \textnormal{atemp}, \textnormal{hum}, and \textnormal{windspeed}; BAQ, \textnormal{PM10}, \textnormal{SO2}, \textnormal{NO2}, \textnormal{CO}, \textnormal{O3}, \textnormal{TEMP}, \textnormal{PRES}, \textnormal{DEWP}, \textnormal{RAIN}, \textnormal{wd}, and \textnormal{WSPM}; HS, \textnormal{PROMO}; and MetroPT-3, \textnormal{TP2}, \textnormal{DV\_\allowbreak pressure}, \textnormal{Motor\_\allowbreak current}, \textnormal{Oil\_\allowbreak temperature}, \textnormal{DV\_\allowbreak eletric}, \textnormal{LPS}, and \textnormal{COMP}.

\subsection{Baselines}
\label{app:baselines}
To ensure a leakage-free and fair evaluation, all models are provided with the same forecast-time available information whenever supported by their input interfaces. For Price, Solar, and Wind, we use the provided future exogenous forecasts. For the remaining datasets, future observed exogenous values are excluded, and only historical exogenous observations and known future calendar features are used. Prophet, iTransformer, TimeXer, TiDE, Chronos-2, Moirai, TimeReasoner, Time-R1, and PostTime support exogenous inputs and receive exactly the same exogenous variables as CastReasoner. For TimeReasoner, Time-R1, and PostTime, we use the prompt templates presented in their respective papers.

\subsection{Experimental Parameters}
\label{app:experimental-parameters}
SFT uses 2,000 trajectories for three epochs with learning rate $2\times10^{-5}$ and batch size 32. GRPO RL uses two epochs with learning rate $10^{-6}$ and batch size 16. The accuracy reward combines nMAE and nMSE with weights 0.9 and 0.1, respectively. The local interval reward weight is 0.2.

All model sizes share the output schema and fixed convergence-guided quotas. RL uses clipping parameter $\epsilon=0.2$, KL coefficient $\beta=0.001$, group size $G=8$, sampling temperature 1.0, maximum completion length 2,048 tokens, and random seed 42.

Trajectory selection uses relative MAE tolerance $\delta=0.03$ and token-rank cap $r_{\max}=100$ for SAD. Cross-domain allocation uses validation weight $\alpha_{\mathrm{val}}=0.9$ and example budget $B=2000$.

For output validity and forecast loss, we use $\lambda_f=1$, $\rho=0.05$, $\epsilon_s=10^{-8}$, $\alpha=0.9$, and $C=4$. Baseline outperformance uses $\lambda_U=0.10$, $\tau_g=0.05$, $c_g=0.20$, $\tau_\Delta=0.02$, $c_\Delta=0.08$, and $\epsilon_U=0.10$. Interval revision uses $\lambda_{\mathrm{loc}}=0.20$, $c_{\mathrm{loc}}=1$, $\lambda_O=0.10$, and $c_O=1$. Teacher-guided reasoning preference uses $\lambda_T=0.02$ and $\tau_T=0.70$.

To compute nMAE and nMSE, we concatenate ground truth from all test prediction windows within each dataset--series group and calculate its mean and population variance. Predictions and ground truth use this shared standardization: each window's MAE is divided by the pooled standard deviation and its MSE by the pooled variance, after which normalized errors are averaged equally across test windows. Pooling avoids inflated normalized errors when an individual window has near-zero variance and provides a common scale across the test set.

\begin{figure*}[t]
\begin{minipage}{\columnwidth}
\centering
\includegraphics[width=\columnwidth]{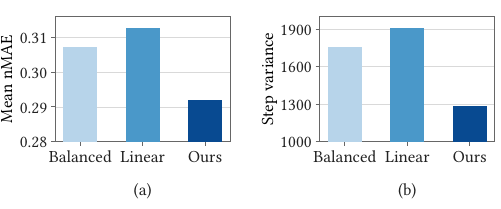}
\caption{Comparison of balanced, linear, and square-root (Ours) data allocation: (a) eight-dataset mean nMAE; (b) variance of best-validation training steps.}
\label{fig:convergence-step-variance}
\end{minipage}\hfill
\begin{minipage}{\columnwidth}
\centering
\includegraphics[width=\columnwidth]{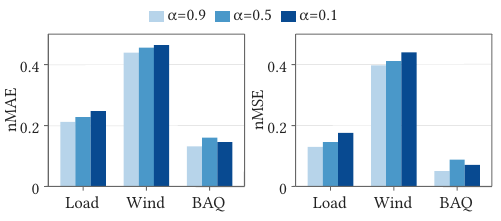}
\caption{Accuracy-reward sensitivity across Load, Wind, and BAQ for $\alpha\in\{0.9,0.5,0.1\}$. Here, $\alpha$ and $1-\alpha$ denote the nMAE and nMSE weights, respectively.}
\label{fig:appendix-reward-balance}
\end{minipage}
\end{figure*}

\section{Additional Experimental Results}
\label{app:additional-results}

\subsection{Cross-Domain Data Allocation}
\label{app:domain-allocation}
We isolate the effect of the allocation rule by fixing the 4B student and total example budget $B$. Balanced allocation assigns equal quotas across domains, whereas our convergence-guided rule uses $p_i\propto\sqrt{T_i}$. For comparison, linear allocation assigns quotas directly in proportion to the pilot convergence steps:
\begin{equation}
  p_i^{\mathrm{lin}}=\frac{T_i}{\sum_{j=1}^{D}T_j},
  \qquad \widetilde{N}_i^{\mathrm{lin}}=Bp_i^{\mathrm{lin}},
\end{equation}
where $D$ is the number of domains and $T_i$ is the pilot training step at which domain $i$ reaches its minimum validation loss under equal sampling. Quotas are fixed from this pilot before formal training. Figure~\ref{fig:convergence-step-variance}(a) shows that square-root allocation has the lowest eight-dataset mean nMAE (0.2919), versus 0.3071 for balanced and 0.3128 for linear allocation. In panel (b), best-validation-step variance rises under linear allocation and falls under square-root allocation relative to balanced sampling. Thus, square-root allocation improves aggregate accuracy while bringing convergence steps closer together across datasets.

\subsection{Model Size}
\label{app:model-size}
Table~\ref{tab:model-size} compares the 0.8B, 4B, and 9B students across eight datasets. Increasing the size from 0.8B to 4B reduces all sixteen dataset--metric errors, with equally averaged nMAE/nMSE decreasing from 0.3346/0.3921 to 0.2919/0.3014. The improvement from 4B to 9B is smaller in aggregate: mean nMAE/nMSE reach 0.2865/0.2877, with gains in nine of the sixteen entries. Both metrics improve on Price, BAQ, and HS, but 4B remains better on Solar and BS. The two metrics also need not favor the same size: on Wind, 9B lowers nMSE from 0.3978 to 0.3821 but raises nMAE from 0.4394 to 0.4443. Similarly, 4B has slightly lower Load nMAE and lower MetroPT-3 nMSE. These results show broad benefits from moving beyond 0.8B, followed by smaller and more dataset-dependent gains at 9B. We use 9B for its stronger overall average, while 4B remains competitive across the evaluated tasks despite using fewer parameters.

\subsection{Accuracy Reward Objective}
\label{app:accuracy-reward-objective}
Figure~\ref{fig:appendix-reward-balance} varies the relative emphasis on nMAE and nMSE in the accuracy reward, weighted by $\alpha$ and $1-\alpha$, respectively. Among the three tested settings, 0.9/0.1 gives the lowest errors on both metrics for Load, Wind, and BAQ. On Load and Wind, both errors rise as more weight shifts to nMSE. BAQ is nonmonotonic: 0.1/0.9 improves on equal weighting but remains worse than 0.9/0.1 on both metrics. Thus, increasing the direct weight on nMSE does not necessarily reduce the final evaluated nMSE. Across these three datasets, the results favor an nMAE-focused reward that retains a smaller nMSE component, supporting the 0.9/0.1 setting used throughout our main experimental comparisons.

\section{Prompt}
\label{app:forecasting-prompt}
Template~\ref{template:forecasting-prompt} supplies the task definition, indexed target history, available context, and cached point forecast $F_0$. Its guidance asks the model to interpret calendar and exogenous signals rather than copy historical windows verbatim. It explicitly distinguishes $F_0$ from the unavailable future targets, clarifying the information boundary during reasoning and strategy selection.

The three strategies define distinct relationships to the prior: retain it unchanged, revise supported aspects using it as the starting point, or reconstruct independently from history and supplied covariates. The reasoning block records the analysis, selected strategy, and forecasting method. The answer block contains exactly $H$ finite values indexed from 0 to $H-1$, with no extra text outside the two blocks. Under retain, every value must exactly match the displayed prior. This structure connects the stated decision to the numerical forecast used in the final evaluation.

\begin{template*}[p]
\centering
\refstepcounter{template}
\label{template:forecasting-prompt}
\begingroup
\definecolor{promptnavy}{RGB}{25,50,120}
\definecolor{promptblue}{RGB}{235,240,255}
\normalfont\rmfamily\normalsize
\newcommand{\promptfence}{\char0\kern0pt\char0\kern0pt\char0}
\begin{tikzpicture}
\node[anchor=north west,outer sep=0pt,inner xsep=10pt,inner ysep=8pt,
      text width=\dimexpr\textwidth-22pt\relax] (promptbody) at (0,-24pt) {%
\normalfont\rmfamily\normalsize
\setlength{\smallskipamount}{1pt}%
\setlength{\parindent}{0pt}%
\setlength{\parskip}{0pt}%
\textbf{Role and objective.} You are forecasting a time series. Use the observed history and supplied future covariates to review the prior forecast $F_0$. Analyze the situational context and forecast prior to infer evidence relevant to forecasting. Choose \textbf{retain}, \textbf{revise}, or \textbf{reconstruct}, and output the complete final forecast in the required format.
\par\smallskip
\textbf{1. Task and visible history.}
\par
\textless forecast\_context\textgreater\\
\{task definition\}\\
\{context information\}\\
\{L observed-window rows, indexed 0 to L-1\}\\
\textless /forecast\_context\textgreater
\par\smallskip
\textbf{Forecasting guidance.} Avoid mechanically copying historical patterns without assessing their relevance to the forecast horizon. When using an analogous historical shape, adjust it using exogenous evidence and context, including future level, peaks, troughs, and timing. Consider holidays, special dates, weekday/weekend changes, and seasonality.
\par\smallskip
\textbf{Field conventions.} Each window is indexed independently from 0; time uses YYYY-MM-DD HH:MM. Covariates include calendar features and available exogenous variables.
\par\smallskip
\textbf{2. Prior forecast and context information.}
\par
\{context information\}\\
\{H forecast-window rows, indexed 0 to H-1 with forecast prior\}
\par\smallskip
\textbf{prior\_forecast} is a cached point forecast $F_0$, not the future target; it is displayed at the observed target precision. Future target values are unavailable.
\par\smallskip
\textbf{3. Decision and final forecast.} 
Choose one option:
\begin{list}{\textbullet}{\setlength{\leftmargin}{12pt}\setlength{\itemsep}{0pt}\setlength{\parsep}{0pt}\setlength{\topsep}{2pt}}
\item \textbf{retain:} Keep $F_0$ unchanged.
\item \textbf{revise:} Use $F_0$ as the starting forecast and correct its level, shape, or timing where the observed history and supplied covariates support a correction.
\item \textbf{reconstruct:} Discard $F_0$ as the forecasting basis and independently construct a new forecast from the observed history and supplied covariates. Do not use $F_0$ as the starting forecast or merely apply corrections to it.
\end{list}
Present your analysis, selected strategy, and forecasting method using the supplied information. Return \mbox{exactly these two blocks:}
\par
\textless think\textgreater\\
\textbf{analysis:} [analyze the situational context and forecast prior to infer evidence relevant to forecasting]\\
\textbf{decision:} [\textbf{retain}, \textbf{revise}, or \textbf{reconstruct}]\\
\textbf{forecast\_method:} [none for \textbf{retain}; corrections to $F_0$ for \textbf{revise}; independent forecast construction for \textbf{reconstruct}]\\
\textless /think\textgreater\\
\textless answer\textgreater\\
\promptfence\\
index \textbar{} value\\
0 \textbar{} [finite numeric value]\\
1 \textbar{} [finite numeric value]\\
...\\
\promptfence\\
\textless /answer\textgreater
\par\smallskip
Output all $H$ finite numeric final forecast values with consecutive indices from 0 to $H-1$. Use sufficient numeric precision for the series scale; decimal places are not fixed. For \textbf{retain}, copy the displayed \textbf{prior\_forecast} values exactly. Replace all output placeholders and the ellipsis with the complete table. Do not output anything outside these two blocks.
};
\coordinate (prompttopright) at ($(promptbody.north east)+(0,24pt)$);
\begin{pgfonlayer}{background}
\fill[promptblue,rounded corners=5pt] (0,0) rectangle (promptbody.south east);
\fill[promptnavy] (0,-24pt) -- (0,-5pt)
  arc[start angle=180,end angle=90,radius=5pt]
  -- ($(prompttopright)+(-5pt,0)$)
  arc[start angle=90,end angle=0,radius=5pt]
  -- (promptbody.north east) -- cycle;
\end{pgfonlayer}
\draw[promptnavy,line width=0.8pt,rounded corners=5pt]
  (0,0) rectangle (promptbody.south east);
\node[anchor=west,inner sep=0pt,text=white,
      font=\normalfont\rmfamily\fontsize{11}{13}\selectfont\bfseries] at (10pt,-12pt)
  {CastReasoner Forecasting Prompt};
\end{tikzpicture}%
\endgroup

\end{template*}

\end{document}